%% file: main.tex
\documentclass[10pt,twocolumn,letterpaper]{article}

\usepackage[accsupp]{axessibility}
\usepackage{cvpr}              

\input{preamble}

\definecolor{cvprblue}{rgb}{0.21,0.49,0.74}
\usepackage[pagebackref,breaklinks,colorlinks,allcolors=cvprblue]{hyperref}
\def\paperID{XXXX} 
\def\confName{CVPR}
\def\confYear{2026}

\title{ERMoE: Eigen-Reparameterized Mixture-of-Experts for Stable Routing and Interpretable Specialization}

\author{
Anzhe Cheng\thanks{Equal contribution.}\qquad
Shukai Duan\footnotemark[1]\qquad
Shixuan Li\qquad
Chenzhong Yin\qquad
Mingxi Cheng\\
Heng Ping\qquad
Tamoghna Chattopadhyay\qquad
Sophia I.~Thomopoulos\\
Shahin Nazarian\qquad
Paul Thompson\qquad
Paul Bogdan\\
University of Southern California\\
{\tt\small \{anzheche,shukaidu,sli97750,chenzhoy,mingxic\}@usc.edu}\\
{\tt\small \{hping,tchattop,sthomopo,shahin.nazarian,pthomp,pbogdan\}@usc.edu}
}
\begin{document}
\maketitle
\input{sec/0_abstract}
\vspace{-7mm}
\input{sec/1_intro}
\vspace{-3mm}
\input{sec/2_related}
\vspace{-3mm}
\input{sec/3_method}
\vspace{-3mm}
\input{sec/4_experiments}
\vspace{-3mm}
\input{sec/5_conclusion}

\input{sec/6_acknowledgment}
{
    \small
    \bibliographystyle{ieeenat_fullname}
    \bibliography{main}
}

\input{sec/X_suppl}

\end{document}

%% file: preamble.tex
\usepackage{adjustbox}
\usepackage{multirow}
\usepackage[linesnumbered,ruled,vlined]{algorithm2e}
\usepackage{float}

\makeatother

%% file: sec/0_abstract.tex
\begin{abstract}

Mixture-of-Experts (MoE) models expand capacity via sparse expert activation, but routing logits can misalign with expert structure (unstable routing, underutilization) and load imbalance can create stragglers. Auxiliary load-balancing losses reduce disparity but often weaken specialization and downstream accuracy. We propose ERMoE, a sparse MoE transformer that reparameterizes each expert in a learned orthonormal eigenbasis and routes with an Eigenbasis Score (cosine similarity between token features and an expert basis) instead of learned gating logits. By tying assignments to each expert’s representation space, ERMoE stabilizes utilization, improves interpretability, and removes explicit balancing losses and their gradient interference. ERMoE reaches state-of-the-art results on ImageNet and image-text retrieval (COCO, Flickr30K) with flatter expert loads. A 3D MRI variant (ERMoE-\textit{ba}) improves brain age prediction by over 7\% and yields anatomically interpretable expert specializations.

\end{abstract}

%% file: sec/1_intro.tex
\section{Introduction}
\label{sec:intro}
\vspace{-3mm}
Deep learning has emerged as a cornerstone of modern artificial intelligence, delivering striking advances across core vision problems~\cite{chai2021deep, lecun2015deep, hinton2012deep}. A dominant driver of this progress has been a straightforward scaling rule: train larger models on more data to steadily reduce error~\cite{kaplan2020scaling}. Yet the same paradigm carries clear limits: ever-larger dense networks imply escalating compute, memory, and energy budgets that are difficult to reconcile with practical training and deployment constraints~\cite{sevilla2022compute,schwartz2020green}. 
To respond to this tension, Mixture-of-Experts (MoE)~\cite{shazeer2017outrageously} architectures were introduced. By activating only a small subset of experts per input, MoE layers decouple parameter count from per-token FLOPs, preserving throughput while expanding capacity. Despite these potential, MoE systems repeatedly encounter structural drawbacks. 

The most significant issue is the misalignment between the router and the experts. With noisy or imbalanced gating, routers may route tokens to suboptimal experts, leading to incorrect predictions and weakening expert specialization~\cite{zhang2025mixture, krishnamurthy2023improving}. Direct evidence comes from R2-T2~\cite{li2025r2}, which shows that test-time re-routing, without updating experts, recovers accuracy by moving tokens away from the originally chosen experts toward better ones, implying the initial router frequently selects the wrong expert. Moreover, routing fluctuations sometimes occur~\cite{dai2022stablemoe}. Under noisy gates, the same input is being sent to different experts during training, while only one is used at inference, which wastes updates and reflects unstable, error-prone assignment under noisy gates. At the representation level, misaligned routing precipitates representation collapse: tokens cluster around expert centroids and experts become redundant, harming performance and blurring roles~\cite{do2024simsmoe, chi2022representation}.

Beyond the misalignment of router experts, routing collapse further constrains MoE progress: token-choice gates amplify a “rich-get-richer” dynamic, in which a few experts attract disproportionate traffic, starving others and creating straggler bottlenecks where the busiest expert dictates end-to-end latency, while overall capacity remains idle~\cite{zhou2022mixture, liu2024routers}. 
In response, mainstream practice adds an auxiliary load-balancing loss (LBL)~\cite{wang2024auxiliary} or switches to balance-by-construction routers~\cite{zhou2022mixture,sun2024ec}. While these strategies improve utilization, they do not address content alignment and can introduce side effects: large LBL coefficients inject interference gradients that impair task performance. Multiple analyses report that strong balancing fosters over-uniform routing and expert overlap, weakening the very divide-and-conquer specialization that MoE seeks to realize~\cite{guo2025advancing, qiu2025demons}. 

\begin{figure*}[t]
\centering
\includegraphics[width=\linewidth]{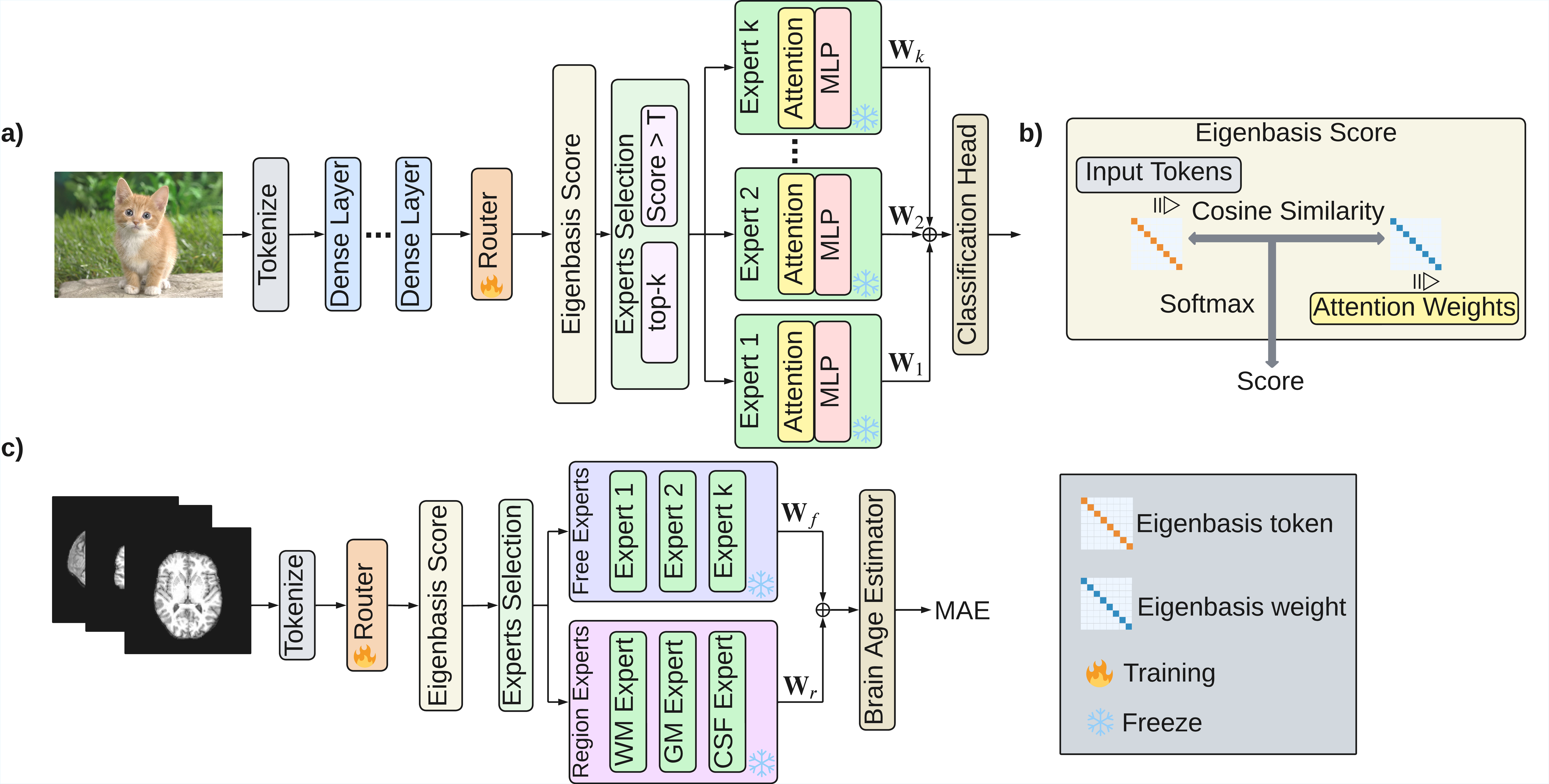}
\caption{\textbf{Overview of the Eigen-Reparameterized Mixture-of-Experts(ERMoE) framework.} a) A ViT backbone tokenizes the image; at each ERMoE block, the router computes an eigenbasis score per expert, selects the top-$k$ experts whose scores exceed a threshold $T$, and aggregates their outputs with score-normalized weights for the classification head.
b) The details of the Eigenbasis score. For a given expert, the input token and its attention-weighted context are projected into that expert’s eigenbasis; the score is the cosine similarity between the two projections.
c) a 3D ViT tokenizes volumetric T1 scans; routing operates over region experts and free experts, and their weighted outputs drive the brain-age estimator.}
\label{fig:ermoe-architecture}
\vspace{-5mm}
\end{figure*} 

To address these issues, we propose the Eigen-Reparameterized Mixture-of-Experts (ERMoE) - a novel MoE whose experts are parameterized in an orthonormal eigenbasis and whose gate scores experts based on feature–basis alignment rather than free router logits. Specifically, each expert’s weight is reparameterized into an eigenbasis, and the router computes the energy captured by each expert’s basis, routing based on these alignment scores. This design ties routing to the experts’ intrinsic representational subspaces, eliminating the need for auxiliary load-balancing losses while directly targeting the root cause of misassignment. By enforcing near-orthogonality within experts, ERMoE reduces feature redundancy and makes expert directions separable, thereby stabilizing specialization and mitigating representation collapse. Using alignment instead of loss-regularized logits improves content-aware assignment without sacrificing sparsity or introducing interference gradients from balancing terms. Additionally, to test whether ERMoE’s eigenbasis routing maintains strong router–expert alignment beyond natural images, we instantiate a 3D neuroimaging variant, ERMoE-\textit{ba}, that scales the 2D ViT backbone in ERMoE to a 3D ViT for brain-age prediction from T1 MRI—an application where brain age is a clinically relevant biomarker.

The primary contributions of this research can be summarized as follows:
\begin{itemize}

\item We introduced Eigen-Reparameterized Mixture-of-Experts(ERMoE),  a novel sparse architecture that parameterizes expert weights in an orthonormal eigenbasis. This fundamentally alters the gating mechanism, replacing standard router logits with a content-aware score based on feature-based alignment.

\item Critically, ERMoE's design resolves the core structural flaws in standard MoEs without compromising sparsity. It eliminates the need for auxiliary load-balancing losses (LBL), thereby avoiding LBL-induced interference gradients and over-uniform expert assignments. 

\item  We proposed ERMoE-$\textit{ba}$, a ERMoE variant specifically designed for 3D neuroimaging. We demonstrate its effectiveness on the challenging task of brain-age prediction from T1 MRI, a clinically relevant application where standard transformers are computationally prohibitive.

\end{itemize}

%% file: sec/2_related.tex
\section{Related Work}
\label{sec:related}
\vspace{-3mm}

\textbf{Vision MoEs.} \textbf{V-MoE}~\cite{riquelme2021scaling} introduced sparse experts to ViT backbones, demonstrating that conditional computation can match or exceed dense models in large-scale image recognition with comparable inference costs. In parallel, \textbf{single-gated MoE} variants~\cite{royer2023revisiting} revisit a simplified training recipe that stabilizes routing via a base branch, asynchronous pipelines, and clustering-based initialization. Beyond transformers, \textbf{DeepMoE}~\cite{wang2020deep} explores deep dynamic routing and per-example channel sparsification inside CNNs, demonstrating that conditional computation benefits extend to convolutional pipelines as well.

\textbf{Routing beyond standard token-choice.} A line of recent work targets the misalignment and instability by modifying the routing rather than the experts. \textbf{Soft MoE}~\cite{puigcerver2023sparse} makes the layer fully differentiable, replacing hard top-$k$ with soft assignments to reduce token drops while maintaining sparse compute at inference. Meanwhile, optimal-transport routers formulate token–expert matching as a balanced assignment, improving utilization and stability in vision MoEs~\cite{vesaghati2024training, liu2024routers}. Another thread, \textbf{BASE layers}~\cite{lewis2021base}, enforces balance by construction via a linear assignment of tokens to experts. While these approaches relieve load imbalance and some routing noise, their scoring often occurs in auxiliary spaces only indirectly tied to each expert’s internal representation.

\textbf{Brain-age prediction.}
For neuroimaging, early deep models framed brain-age estimation from T1-weighted MRI as volumetric regression using compact 3D CNNs~\cite{yin2023anatomically, yin2025deep}. \textbf{SFCN}~\cite{peng2021accurate} established a strong and lightweight 3D CNN baseline for T1-weighted MRI brain-age regression, and transformer-based designs~\cite{he2021global} have since shown competitive accuracy by combining long-range context with local detail. Pretraining and adaptation have also become central: recent work evaluates pretraining 3D architectures on video or synthetic MRI and reports consistent improvements on downstream neuroimaging tasks, including brain age~\cite{dhinagar2023video}. Complementarily, \textbf{PEFT}~\cite{dhinagar2025peft} has been shown to adapt foundation transformers to neuroimaging with far fewer trainable parameters while matching or surpassing full fine-tuning in data-limited regimes. 

%% file: sec/3_method.tex
\section{Method}
\label{sec:method}
\vspace{-3mm}
As shown in Fig.~\ref{fig:ermoe-architecture}a, ERMoE augments a ViT-style backbone with MoE blocks in which each expert is reparameterized by an orthonormal eigenbasis and routing is performed by feature-basis alignment. Given token embeddings, the router computes, for every expert, an eigenbasis score by projecting the current token and its attention-weighted context into that expert’s basis and taking their cosine similarity. Experts whose scores exceed a confidence threshold $T$ are eligible; the router then selects the top-\(k\) among those experts who got $\text{Score}\geq T$. Mixture weights \(w_e\) are obtained by normalizing the positive scores of the selected experts, and the block output is their weighted sum \(\sum_{e} w_e f^{(e)}(x)\), which is passed to the task head. 

\textbf{Eigen-Reparameterized Experts.} Each expert $e\!\in\!\{1,\dots,E\}$, whose weight $\mathbf{W}^{(e)}$ is parameterized in an \emph{orthonormal eigenbasis}:
\begin{equation}
  \mathbf{W}^{(e)} \;=\; \mathbf{U}^{(e)} \,\mathrm{diag}(s^{(e)})\, \mathbf{V}^{(e)\top},
\end{equation}
Here, $\mathbf{U}^{(e)},\mathbf{V}^{(e)}\in\mathbb{R}^{d\times d}$ are orthonormal bases with size of feature dimensionality $d$. Specifically, $\mathbf{U}^{(e)\top}\mathbf{U}^{(e)}=\mathbf{V}^{(e)\top}\mathbf{V}^{(e)}=\mathbf{I}$, and $s^{(e)}\in\mathbb{R}^d$ are the trainable coefficients. In practice, we maintain orthonormality with a light penalty and occasional re-orthogonalization, which yields decorrelated directions that make scores geometrically meaningful.

\textbf{Projection to the Eigenbasis.}
As illustrated in Fig~\ref{fig:ermoe-architecture}b, for each token $i$ we form a local context vector $c_i$ from the self-attention in the current block. Concretely, let $z_i \in \mathbb{R}^d$ denote the output of the attention layer for token $i$, and let $\alpha_{ij}$ be the attention weights from token $i$ to token $j$ in that block. We define an attention-weighted context
\begin{equation}
c_i \;=\; \sum_{j} \alpha_{ij}\, z_j,
\end{equation}
and normalize both the token and its context before projection:
\begin{equation}
\tilde{x}_i \;=\; \frac{x_i}{\|x_i\|_2}, \qquad \tilde{c}_i \;=\; \frac{c_i}{\|c_i\|_2}.
\end{equation}
For each expert $e$, we then project these normalized vectors into that expert's eigenbasis:
\begin{equation}
u_i^{(e)} = U^{(e)\top} \tilde{x}_i, \qquad v_i^{(e)} = U^{(e)\top} \tilde{c}_i.
\label{eq:proj}
\end{equation}

\textbf{Eigenbasis Score.}
The $\mathrm{Score}$ is the cosine similarity between the two projections in Eq.~\ref{eq:proj}:
\begin{equation}
  \mathrm{Score}_e(i) 
  \;=\; \cos\big(u_i^{(e)},\, v_i^{(e)}\big) 
  \;=\; \frac{\langle u_i^{(e)},\, v_i^{(e)}\rangle}{\|u_i^{(e)}\|_2\,\|v_i^{(e)}\|_2}.
  \label{eq:score}
\end{equation}
$\mathrm{Score}_e(i) \in [-1,1]$ measures how well token $i$ and its attention context \emph{align} along expert $e$'s basis directions. Aligning routing with this geometry makes selection content-aware while being robust to logit noise.

\textbf{Confidence Threshold and Top-\texorpdfstring{$k$}{k} Routing.}
We introduce a confidence threshold $T\!\in\![0,1)$ to suppress low-confidence routes:
\begin{equation}
  \mathcal{S}_i 
  \;=\; \big\{\, e \;:\; \mathrm{Score}_e(i)\ge T \,\big\}, .
\end{equation}
We denote $\mathcal{S}_i^{(k)}$ as Top-\texorpdfstring{$k$}{k} experts in $\mathcal{S}_i$ by $\mathrm{Score}_e(i)$. If all $|\mathcal{S}_i|\!<\!k$, we fall back to the highest-scoring experts to fill $\mathcal{S}_i^{(k)}$. The threshold $T$ filters spurious alignments, stabilizes utilization, and reduces misrouting by enforcing a minimal geometric match before activation. In practice, $T$ is chosen on validation to meet a target activation budget and minimize token drops at a fixed capacity (Details of choosing threshold $T$ are in the Appendix).

\textbf{Mixture Weights and Expert Fusion.}
Given the selected experts $\mathcal{S}_i^{(k)}$, we convert scores to nonnegative mixture weights and fuse expert outputs. Let $f^{(e)}(\cdot)$ be expert $e$’s function (e.g., MLP). Define
\begin{equation}
  w_e(i) \;=\; \frac{\max\{\,\mathrm{Score}_e(i),\,0\,\}}{\sum\limits_{e'\in \mathcal{S}_i^{(k)}} \max\{\,\mathrm{Score}_{e'}(i),\,0\,\}} \;\;\; \text{for } e\in \mathcal{S}_i^{(k)},
  \label{eq:weights}
\end{equation}
where $\sum_{e\in \mathcal{S}_i^{(k)}} w_e(i)=1$, and the block’s routed output for token $i$ is the weighted sum
\begin{equation}
  y_i \;=\; \sum_{e\in \mathcal{S}_i^{(k)}} w_e(i)\, f^{(e)}(x_i).
  \label{eq:fusion}
\end{equation}

In our implementation, the per-expert weight $w_e$ is derived from the eigenbasis score and used to scale each expert’s contribution before aggregation. This preserves strict sparsity (top-$k$) and provides a smooth interpolation among the selected experts.

\textbf{Loss function.}
We optimize the task loss with a light orthogonality regularizer on each expert’s basis. For expert $e\in{1,\dots,E}$, the orthogonality penalty is
\begin{equation}
    \mathcal{L}^{(e)}_{\text{ortho}} = \big\|\mathbf{U}^{(e)\top}\mathbf{U}^{(e)}-\mathbf{I}\big\|_F^2
    +  \big\|\mathbf{V}^{(e)\top}\mathbf{V}^{(e)}-\mathbf{I}\big\|_F^2,
\end{equation}

where $\big\|\cdot\big\|_{F}$ denotes the Frobenius norm~\cite{bottcher2008frobenius} and $\mathbf{I}$ is the identity of matching size. The total objective is

\begin{equation}
  \mathcal{L}
  \;=\;
  \mathcal{L}_{\text{task}}
  \;+\;
  \lambda \sum_{e=1}^{E}\!\mathcal{L}^{(e)}_{\text{ortho}},
  \label{eq:loss}
\end{equation}

with $\lambda=5\times10^{-5}$(Details on selecting $\lambda$ in our experiment are provided in the Appendix.). Here $\mathcal{L}_{\text{task}}$ is the standard task loss (e.g., cross-entropy for classification). We do not include any auxiliary load-balancing term; usage balance arises from geometric routing with the thresholded top-$k$.

\textbf{3D Adaptation for Brain MRI.} To evaluate router-expert alignment, we further adapted our ERMoE to 3D brain images to form ERMoE-\textit{ba}. For volumetric T1 MRI $X\!\in\!\mathbb{R}^{H\times W\times D\times C}$, we adopt 3D ViT-style patching: $X$ is partitioned into non-overlapping $P^3$ cubes, flattened and linearly projected to embeddings with 3D positional encodings. ERMoE blocks replace FFNs in the encoder; attention and patch embeddings follow standard 3D Transformer practice. 

We then designed two expert types: \textbf{region experts} and \textbf{free experts}. 
As shown in Fig.~\ref{fig:ermoe-architecture}c, region experts are pre-trained on parcellations of the brain, i.e., one expert only trained on White Matter(WM), one only trained on Grey Matter(GM), and another trained on Cerebrospinal fluid(CSF) only. 
Free experts receive whole-brain patches without regional restriction and discover complementary subspaces. 

From the final encoder stage, we obtain a global representation $h\!\in\!\mathbb{R}^{d}$ (via CLS token). ERMoE-$\mathit{ba}$ uses a classification head over discretized age bins $\{a_1,\dots,a_B\}$ with logits $g\in\mathbb{R}^{B}$, and predicts age by the expected value
\begin{equation}
  p_b \;=\; \mathrm{softmax}\!\left(\frac{g_b}{\tau}\right), 
  \qquad 
  \hat{\mathrm{Age}} \;=\; \sum_{b=1}^{B} a_b\, p_b,
  \label{eq:age_head}
\end{equation}
optionally with temperature $\tau$ for calibration. This “classification-then-expectation” formulation is a standard, robust alternative to direct regression in biological age estimation.

%% file: sec/4_experiments.tex
\section{Experiments}
\label{sec:exp}
\vspace{-2mm}
To validate the effectiveness of our ERMoE, we conducted comprehensive experiments on both natural image datasets and 3D brain MRI dataset. To show the load balance of each expert, we also analyzed the activity of experts during each training and inference stage.
\vspace{-2mm}
\subsection{Experiment setup}
\label{sec:exp_setup}
\vspace{-1mm}
\textbf{Datasets.} The \textbf{CIFAR-10} and \textbf{CIFAR-100} are obtained from the TensorFlow datasets~\cite{abadi2016tensorflow}. CIFAR-10~\cite{krizhevsky2009learning} consists of 60,000 images, each of size $32\times32$.
CIFAR-100~\cite{krizhevsky2009learning} has the same image size and total count but spans 100 classes, organized into 20 superclasses with both coarse and fine labels.
\textbf{Tiny ImageNet}~\cite{le2015tiny} consists of a dataset of $100,000$ images distributed across 200 classes, with 500 images per class for training, and an additional set of $10,000$ images for testing.
\textbf{ImageNet-1K}~\cite{russakovsky2015imagenet} includes 1{,}281{,}167 training images, 50{,}000 validation images, and 100{,}000 test images over 1{,}000 categories.
\textbf{MS COCO}~\cite{lin2014microsoft} contains everyday scenes with rich instance-level annotations for detection, segmentation, and captioning, with 113{,}287 training images and 5{,}000/5{,}000 images for validation/test.
\textbf{Flickr30k}~\cite{plummer2015flickr30k} comprises 31{,}783 images, each paired with five human-written captions.
Alzheimer’s Disease Neuroimaging Initiative (\textbf{ADNI})~\cite{petersen2010alzheimer} is a longitudinal, multi-center observational study designed to validate imaging and fluid biomarkers for Alzheimer’s disease. We analyzed 7{,}965 T1-weighted MRI scans, parcellated with FreeSurfer~\cite{fischl2012freesurfer} following the ENIGMA FreeSurfer protocol~\cite{enigmaFSprotocol,enigmaFSgithub}, and performed quality control in line with ENIGMA recommendations~\cite{grasby2020}.

\textbf{Backbones.}
For 2D images, we adopt ViT backbones with pretrained ViT-B weights and insert ERMoE blocks in place of FFNs.
For 3D MRI, we adopt a 3D ViT: volumes are tokenized into non-overlapping $16^3$ cubes with 3D positional encodings; ERMoE blocks replace FFNs analogously. 

\begin{table}[h!]
\vspace{-3mm}
\centering
\begin{tabular}{lccc}
\toprule
\textbf{Method} & \textbf{Top-1 (\%)} &\textbf{Top-5 (\%)} \\
\midrule
V\!-\!MoE   & \underline{87.41} & \underline{97.94} \\
Single-gated MoE   & 72.38 & 93.26 \\
DeepMoE   & 77.12 & 95.07  \\
\midrule
\textbf{ERMoE} & \textbf{88.03} & \textbf{98.97}   \\
\bottomrule
\end{tabular}
\caption{Top-1 and Top-5 classification accuracy(\%) with different methods on \textbf{ImageNet}. \textbf{Bold} denotes the best results and \underline{underline} the second‐best results.}
\label{tab:imagenet_compare}
\vspace{-3mm}
\end{table}

\begin{table*}[t]
\centering
\begin{adjustbox}{max width=\linewidth,center}
  \setlength{\tabcolsep}{5pt}
  \renewcommand{\arraystretch}{0.95}
  \begin{tabular}{lcccccc}
    \toprule
    & \multicolumn{2}{c}{\textbf{CIFAR-10}} & \multicolumn{2}{c}{\textbf{CIFAR-100}} & \multicolumn{2}{c}{\textbf{Tiny-ImageNet-200}} \\
    \cmidrule(lr){2-3}\cmidrule(lr){4-5}\cmidrule(lr){6-7}
    \textbf{Method} & 5-shot & 10-shot & 5-shot & 10-shot & 5-shot & 10-shot \\
    \midrule
    V\!-\!MoE             & \underline{94.62} & \textbf{98.94} & \underline{89.49} & \underline{91.26} & \underline{81.12} & \underline{83.16} \\
    Single-gated MoE      & 90.30 & 94.79 & 74.85 & 76.62 & 60.45 & 62.49 \\
    DeepMoE               & 90.28 & 93.46 & 68.81 & 70.58 & 55.46 & 57.49 \\
    \midrule
    \textbf{ERMoE}                 & \textbf{96.05} & \underline{98.79} & \textbf{93.47} & \textbf{96.94} & \textbf{83.17} & \textbf{85.04} \\
    \bottomrule
  \end{tabular}
\end{adjustbox}
\caption{Top-1 (\%) for 5/10-shot linear probes on CIFAR-10, CIFAR-100 and Tiny-ImageNet-200. \textbf{Bold} denotes the best results and \underline{underline} the second‐best results.}
\label{tab:fewshot_combined}
\vspace{-5mm}
\end{table*}

\textbf{Training Settings.}
Unless otherwise stated, each MoE layer has $E{=}8$ experts (For 3D MRI, each layer contains 3 region experts and 5 free experts) and routes top-$k{=}2$ experts per token with a threshold set to 0.5.
We follow modern ViT recipes: AdamW optimizer, cosine annealing with warmup, and strong augmentations.
Unless noted, training uses batch size 100, base learning rate scaled linearly with batch size ($\text{lr} = 1\text{e-}4 \times \tfrac{\text{bs}}{256}$), weight decay $0.05$, warmup 5 epochs, cosine schedule to zero, label-smoothing $0.1$.Training epochs are 300 for fine-tuning on ImageNet-1K and ADNI, 5/10 epochs for CIFAR and Tiny-ImageNet for few-shot learning.

\subsection{Classification Results on Images}
We first evaluated our model's ability to improve feature representations for natural image classification when integrated into a standard ViT backbone. To demonstrate our model's generalizability, we trained and validated it on a large, complex dataset - ImageNet - and assessed few-shot linear probes on smaller datasets, including CIFAR and Tiny-ImageNet. These will help us isolate the contribution of our eigenbasis-aligned routing with our novel expert selection technique. 

\textbf{Classification on ImageNet.}
Our model is trained and evaluated on the challenging ImageNet-1K benchmark to obtain large-scale classification performance. We used pretrained ViT-B/16\footnote{Pretrained weights from \texttt{torchvision.models.vit\_b\_16}. We also reference the official ViT-B/16 checkpoints released by Google Research (ImageNet-21k pretrain + ImageNet-1K fine-tune).} backbones, replacing the standard FFN layers at identical block indices with the respective MoE blocks. This substitution ensures a fair, controlled comparison of the routing strategies, isolating the impact of the gating mechanism and expert parameterization. All models were fine-tuned from the same checkpoint under identical training conditions, as detailed in Section~\ref{sec:exp_setup}.

The classification results are presented in Tab.~\ref{tab:imagenet_compare}. Our ERMoE achieves more than 88\% Top-1 accuracy and nearly perfect Top-5 accuracy. This surpasses the strong V-MoE baseline by about 1\% on both metrics. The improvement is not huge, but significant, as the ImageNet classification task already sets a very high standard for now. Compared to other vision MoEs, the key structural difference is that ERMoE achieves this result without any auxiliary Load-Balancing Loss(LBL). In contrast, V-MoE relies heavily on an LBL to prevent routing collapse. Our superior performance indicates that the LBL, while necessary for stabilizing standard MoEs, acts as a performance bottleneck. ERMoE's alignment-based routing, by intrinsically tying token assignment to the experts' representational subspaces, promotes natural specialization and stable utilization, thus avoiding this fundamental accuracy-stability trade-off.

Furthermore, the stark performance degradation of other baselines, such as Single-gated MoE and DeepMoE, highlights the critical nature of the routing problem. These methods, lacking a robust balancing mechanism, are highly susceptible to the phenomena of routing and representation collapse.

\textbf{Few-shot classification on CIFAR and Tiny-ImageNet.}
Beyond performing our model on a single dataset, we tested our model on small and medium-scale datasets. We utilized our pretrained model on ImageNet and conducted a few-shot classification evaluation. Following the standard linear probe protocol~\cite{kolesnikov2019revisiting}, we freeze the entire backbone and train only a new linear classifier head for 5 and 10 epochs. This low-data regime is highly sensitive to the linear separability and generalizability of the learned features; models suffering from feature redundancy or representation collapse, a known issue in standard MoEs, are expected to perform poorly.

As shown in Tab.~\ref{tab:fewshot_combined}, our ERMoE demonstrates a commanding lead over all baselines across all datasets. On the fine-grained CIFAR-100 5-shot task, ERMoE achieves \textbf{93.47\%} accuracy, surpassing the strong V-MoE baseline (89.49\%) by a significant margin of almost 4\%. This gap is even more pronounced on the more challenging Tiny-ImageNet dataset, where ERMoE outperforms V-MoE by a massive \textbf{7\%}. The performance of other baselines, such as Single-gated MoE (74.85\% on CIFAR-100) and DeepMoE (68.81\% on CIFAR-100), is substantially lower, indicating that their routing or specialization strategies fail to produce effective representations.


\vspace{-2mm}
\subsection{Cross-Modal Image-Text Retrieval}
\vspace{-2mm}
To verify that ERMoE’s expert-diverse representations extend beyond classification to general multimodal understanding, we evaluate it as the vision encoder in image–text retrieval—a setting where MoE has proved well-suited for multimodal learning.

\textbf{Setup.} We adopt a CLIP-based retrieval framework and replace CLIP’s original ViT-B/16 image encoder with our ERMoE vision transformer (pretrained on ImageNet-1K), forming a model we call CLIP-ERMoE. To isolate the contribution of the image representations, the text tower remains frozen with its pretrained CLIP weights (i.e., we do not fine-tune or modify the text encoder). We compared CLIP-ERMoE against several strong vision-language baselines, including Long-CLIP~\cite{zhang2024long}, Fix-CLIP~\cite{wang2025fix}, and CLIP-MoE~\cite{zhang2024clip}.

\begin{table}[ht]
\centering
\small
\setlength{\tabcolsep}{4.2pt}
\begin{tabular}{lcccccc}
\toprule
& \multicolumn{3}{c}{\textbf{COCO I2T}} & \multicolumn{3}{c}{\textbf{COCO T2I}} \\
\cmidrule(lr){2-4}\cmidrule(lr){5-7}
Model & @1 & @5 & @10 & @1 & @5 & @10 \\
\midrule
Long-CLIP   & 62.8 & 85.1 & 91.2 & 46.3 & 70.8 & 79.8 \\
Fix-CLIP    & 62.3 & 85.4 & 91.4 & \textbf{49.1} & \textbf{73.8} & \textbf{82.4} \\
CLIP-MoE    & \underline{65.0} & \underline{86.0} & \underline{92.0} & \underline{46.8} & 71.7 & 80.4 \\
\midrule
\textbf{CLIP-ERMoE}  & \textbf{65.4} & \textbf{88.3} & \textbf{94.7} & 46.5 & \underline{72.9} & \underline{81.6} \\
\bottomrule
\end{tabular}
\caption{COCO: Image-to-Text (I2T) and Text-to-Image (T2I) Recall@\{1,5,10\}. \textbf{Bold} denotes the best results and \underline{underline} the second‐best results.}
\label{tab:coco-retrieval}
\vspace{-3mm}
\end{table}

\textbf{Retrieval Performance.} Tab.~\ref{tab:coco-retrieval} and Tab.~\ref{tab:flickr-retrieval} present the image-text retrieval results on MS-COCO and Flickr30K, respectively. Focusing first on the image-to-text (I2T) results, CLIP-ERMoE achieves the highest recall rates on both datasets. On COCO, our model achieves a state-of-the-art I2T R@1 of 65.4\%, and outperforms other methods by around 2\%. The advantage of CLIP-ERMoE grows at higher recall ranks (e.g., 94.7\% at R@10, which is almost 3\% above the second-best). A similar trend is observed on Flickr30K: CLIP-ERMoE delivers 63.4\% I2T R@1, comfortably surpassing the strongest dense model (Fix-CLIP at 60.5\%) and the prior MoE-based model (CLIP-MoE at 60.5\%). It also essentially matches or exceeds the best competing results with threshold decreases. These improvements in I2T retrieval indicate that the expert-diverse image features learned by ERMoE are well-suited for fine-grained cross-modal alignment.

\begin{table}[ht]
\centering
\small
\setlength{\tabcolsep}{4.2pt}
\begin{tabular}{lcccccc}
\toprule
& \multicolumn{3}{c}{\textbf{Flickr I2T}} & \multicolumn{3}{c}{\textbf{Flickr T2I}} \\
\cmidrule(lr){2-4}\cmidrule(lr){5-7}
Model & @1 & @5 & @10 & @1 & @5 & @10 \\
\midrule
Long-CLIP   & 53.4 & 77.5 & 85.3 & 41.2 & 64.1 & 72.6 \\
Fix-CLIP    & \underline{60.5} & \underline{89.0} & \textbf{89.8} & \textbf{49.6} & \textbf{66.9} & \textbf{77.4} \\
CLIP-MoE    & \underline{60.5} & 82.3 & \underline{88.8} & 42.1 & 64.7 & 73.2 \\
\midrule
\textbf{CLIP-ERMoE}  & \textbf{63.4} & \textbf{89.3} & \textbf{89.8} & \underline{42.4} & \underline{64.9} & \underline{74.0} \\
\bottomrule
\end{tabular}
\caption{Flickr30k: Image-to-Text (I2T) and Text-to-Image (T2I) Recall@\{1,5,10\}.\textbf{Bold} denotes the best results and \underline{underline} the second‐best results.}
\label{tab:flickr-retrieval}
\vspace{-3mm}
\end{table}

For text-to-image (T2I) retrieval, the overall pattern is different. No method in our comparison achieves a new state-of-the-art on T2I, and CLIP-ERMoE’s performance, while competitive, does not surpass the current best. Our CLIP-ERMoE ranks second on COCO T2I (with 81.6\% R@10, slightly below Fix-CLIP’s 82.4\%) and is on par with CLIP-MoE on Flickr T2I (around 42\% R@1 and 74\% R@10). We attribute the fact that all models’ T2I scores are relatively lower and not SOTA to the fixed text encoder in our setup. Because we froze the CLIP text tower and made no modifications to adapt it, the model’s ability to handle diverse text queries is constrained by the original CLIP text representations. In retrieval, the text-to-image direction is particularly sensitive to how well the text embeddings can differentiate fine nuances and align with image features. Methods like Fix-CLIP likely benefit from additional text-encoder training that enhances the textual side of the alignment, giving them an edge in T2I. By contrast, our approach prioritized improving the image encoder while leaving the language encoder untouched; as a result, the full potential of ERMoE’s image features isn’t realized on T2I queries, since the text queries are still represented in the same old CLIP space without refinement.

Despite this limitation on the T2I side, the strong I2T results of CLIP-ERMoE confirm that our expert-rich image representations generalize well to multimodal retrieval tasks. Without any custom pretraining for image-text alignment, the ImageNet-pretrained ERMoE encoder, when plugged into the CLIP framework, is able to outperform dedicated multimodal models in finding relevant captions.

\vspace{-2mm}
\subsection{Results on 3D brain MRI}
\vspace{-1mm}

Beyond natural images, we scale ERMoE to volumetric neuroimaging by applying it to 3D brain MRI— a substantially higher-dimensional setting—using the longitudinal ADNI cohort as our benchmark. Consistent with prior evidence that transformer encoders handle 3D medical volumes effectively, these experiments demonstrate that ERMoE’s expert-diverse features transfer beyond 2D photos to high-dimensional neuroimaging.

\textbf{Dataset division.} We partition the ADNI cohort subject-wise into training, validation, and test sets in an 8:1:1 ratio. All parameters, including expert modules, are fit solely on the training set; validation and test data remain unseen during training. We validate at each epoch and, after training, evaluate on the held-out test set. We report MAE on the validation and test sets.

\begin{table}[ht]
\centering
\small
\setlength{\tabcolsep}{3pt}
\renewcommand{\arraystretch}{1.05}
\begin{tabular}{lccc}
\toprule
\textbf{Model} & \textbf{Male} & \textbf{Female} & \textbf{Total} \\
\midrule 
3D CNN~\cite{peng2021accurate}            & 3.03 & 3.41 & 3.13 \\
3D ViT~\cite{zhang2024triamese}           & 3.49 & 3.55 & 3.52 \\
3D Swin Transformer~\cite{kim2024domain}  & 2.84 & 2.81 & 2.83 \\
\midrule
\textbf{ERMoE-\textit{ba}}                & \textbf{2.30} & \textbf{2.32} & \textbf{2.31} \\
\bottomrule
\end{tabular}
\caption{\textbf{Brain-age (BA) on ADNI test set.}
Mean absolute error (MAE$\downarrow$, years) for male, female, and all subjects.
Baselines are representative families (3D CNN~\cite{peng2021accurate}, 3D ViT~\cite{zhang2024triamese}, 3D Swin Transformer~\cite{kim2024domain}). 
All numbers were reproduced under our pipeline on the held-out ADNI test set.}
\label{tab:adni_ba_mae}
\vspace{-3mm}
\end{table}

\textbf{Performance on ADNI.} On the held-out ADNI test set, \textbf{ERMoE-\textit{ba}} attains a total MAE of \textbf{2.31} years, improving over strong, representative families reproduced under our pipeline—3D CNNs (SFCN-style: \textbf{3.13}), 3D ViT (\textbf{3.52}), and 3D Swin Transformer (\textbf{2.83})—by \textbf{0.82} (\textbf{26.2\%}), \textbf{1.21} (\textbf{34.4\%}), and \textbf{0.52} years (\textbf{18.4\%}), respectively (Table~\ref{tab:adni_ba_mae}). These gains hold across architecture families widely used for brain-age estimation—lightweight 3D CNNs such as SFCN, ViT-style transformers, and 3D Swin variants—supporting the view that the benefit comes from routing rather than a particular backbone. Sex-stratified MAE further shows that \textbf{ERMoE-\textit{ba}} improves both groups, reaching \textbf{2.30} years for males and \textbf{2.32} years for females, which also yields superior results compared to other baselines. 

\begin{figure}[t]
  \centering
  \includegraphics[width=\linewidth]{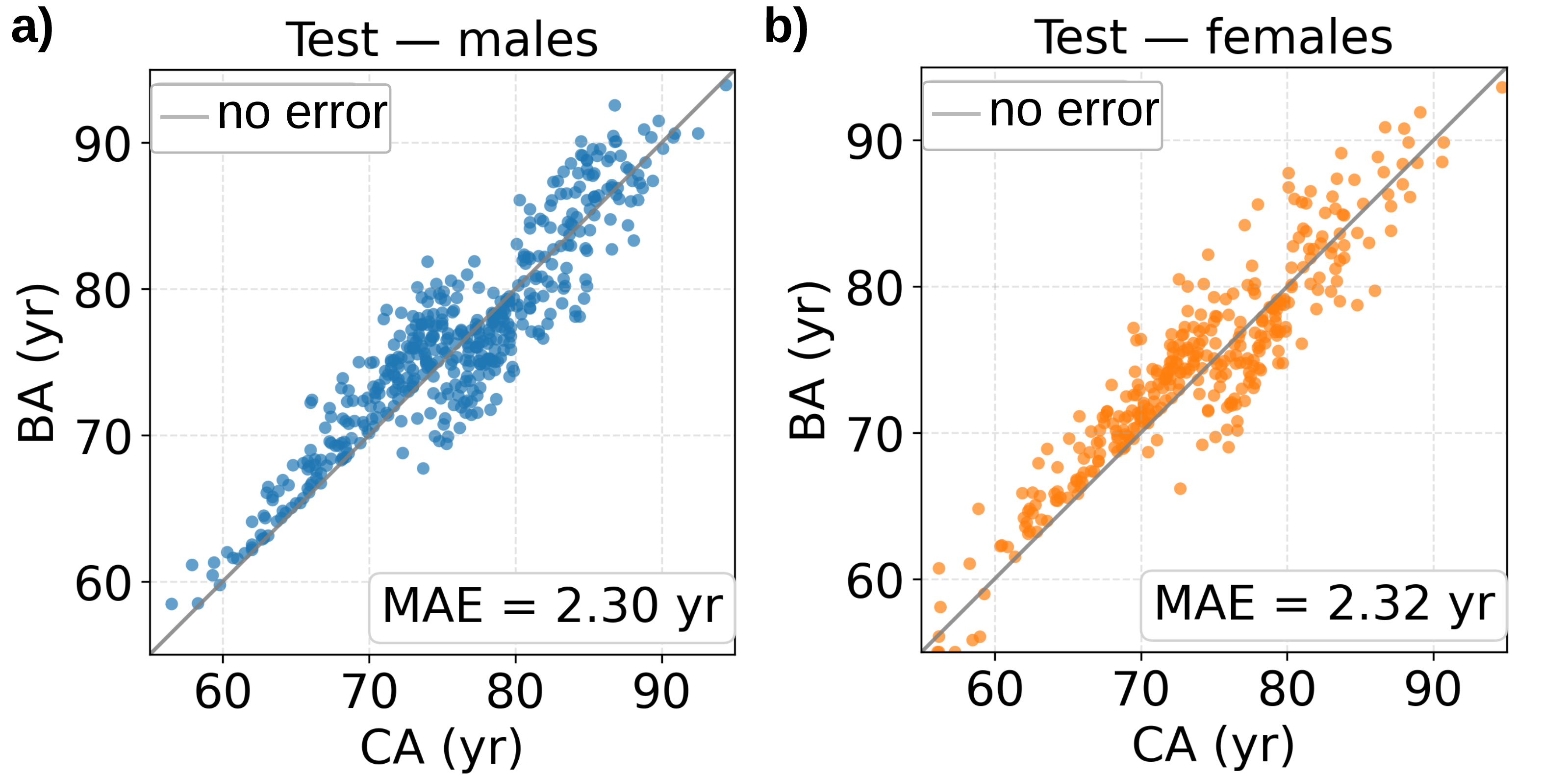}
  \caption{\textbf{Brain-age (BA) estimation on ADNI test set.}
Scatter plots show ERMoE-\textit{ba} predicted BA versus chronological age (CA) for a) males and b) females on the test set. Points are colored by sex (male: blue; female: orange). The solid diagonal denotes the ``no error'' line ($\mathrm{BA}=\mathrm{CA}$).}
  \label{fig:brain_mae}
  \vspace{-7mm}
\end{figure}

The sex-stratified BA–CA plots (Fig.~\ref{fig:brain_mae}) show why ERMoE-\textit{ba} tightens the relation around the identity line across the full age range and suppresses the large residuals that typically appear at the youngest/oldest ages. Best-fit trends approach a unit slope with a near-zero intercept, indicating improved calibration—an essential requirement for utilizing brain age as a biomarker. These patterns align with the mechanism of ERMoE-\textit{ba}: tokens are routed to experts whose eigenbases are aligned with age-sensitive morphometry (e.g., ventricular expansion, cortical thinning), thereby reducing misassignment and representation overlap without auxiliary load-balancing losses that can introduce interference gradients.

Crucially, the male/female panels exhibit comparable dispersion, suggesting that ERMoE’s content-aware, thresholded top-$k$ routing improves accuracy and calibration without introducing sex-specific degradation. In summary, tying the router to experts’ representational subspaces results in lower error and better-conditioned BA–CA relationships on ADNI than CNN, ViT, or Swin-based baselines, while preserving sparse computation.

\vspace{-3mm}
\subsection{Experts Activities}
\vspace{-2mm}
To assess the reliability of ERMoE’s routing design, we analyze how the router distributes tokens across experts during training and inference. This includes examining whether tokens are consistently assigned to the most appropriate experts (i.e., functional specialization) and whether expert loads are well balanced to prevent collapse or bottlenecks. The following analyses investigate these dynamics from both representational and systems perspectives.

\begin{figure}[t]
  \centering
  \includegraphics[width=0.9\linewidth]{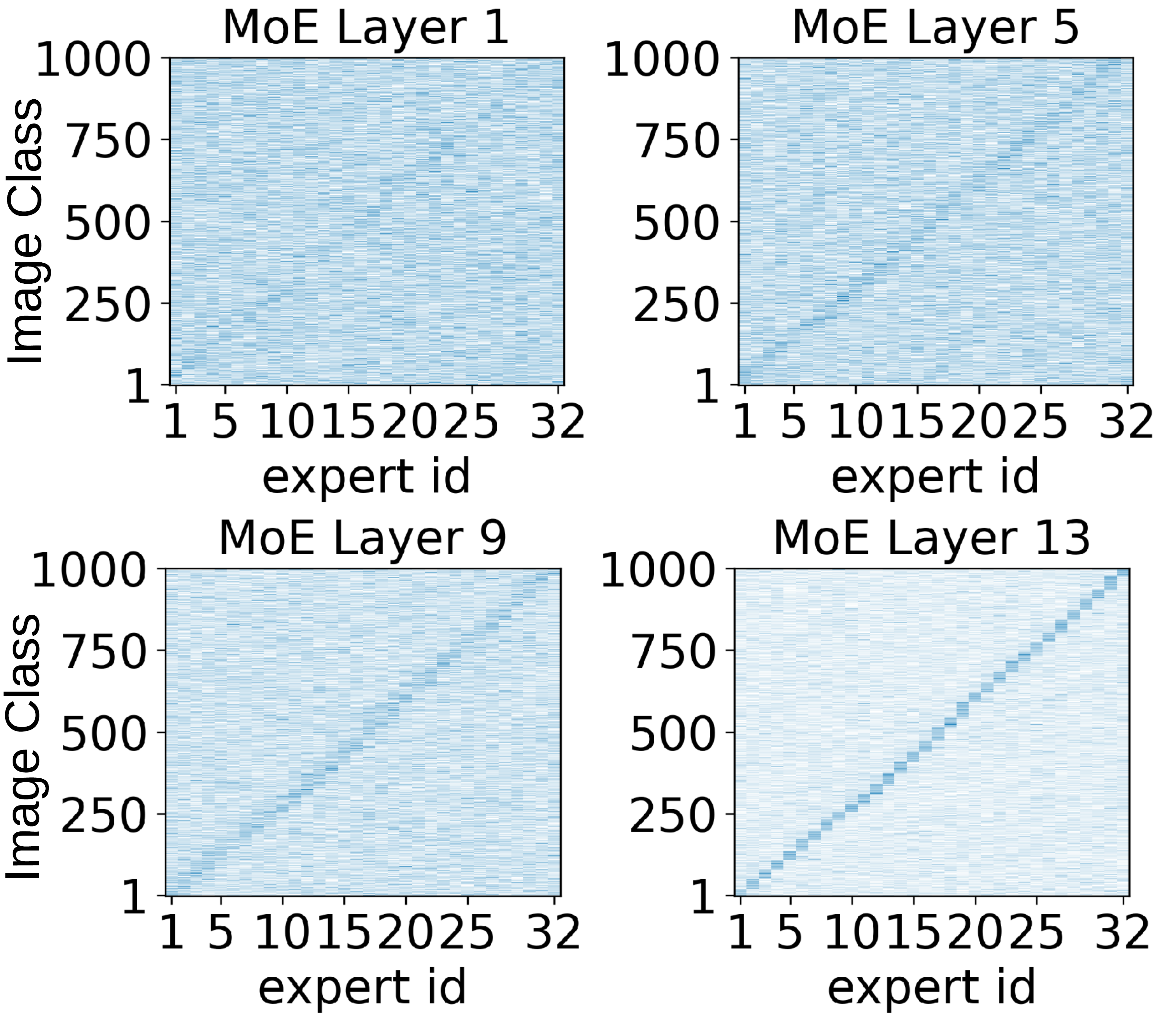}
  \caption{\textbf{Experts activation during training (ERMoE).}
  Class–expert routing heatmaps for four ERMoE layers in a ViT-B/16 backbone. 
  Each panel shows the average mixture weight assigned to each expert (x-axis: expert id; y-axis: ImageNet class index $0\!\rightarrow\!1000$). 
  Early layers display broad, overlapping activations; deeper layers sharpen into clearer preferences, indicating healthy specialization without collapse. 
  ERMoE uses thresholded top-$k$ routing with eigenbasis-aligned scores, which curbs noisy assignments and maintains balanced utilization throughout training.}
  \label{fig:experts-train}
  \vspace{-5mm}
\end{figure}

\textbf{Experts Activation during Training.} Fig.~\ref{fig:experts-train} visualizes ERMoE’s routing dynamics across four MoE layers during ImageNet training by plotting, for each layer, the average per-class mixture weight assigned to each expert. Early in training, the activations are diffuse—multiple experts share traffic for most classes—while deeper layers and later epochs exhibit sharper, yet still imperfect, class–expert preferences. This non-orthogonal pattern is desirable: experts specialize without collapsing to a one-to-one class partition, consistent with observations that deeper sparse ViT layers correlate more strongly with semantic structure but retain overlap~\cite{riquelme2021scaling}. Importantly, ERMoE’s eigenbasis-aligned, thresholded top-$k$ gate suppresses noisy routes that commonly arise in token-choice MoEs, reducing the routing fluctuation, while achieving balanced utilization without auxiliary load-balancing losses. Moreover, ERMoE ties routing scores directly to each expert’s representational subspace, which yields stable specialization over training and avoids expert collapse or straggler bottlenecks typically caused by rich-get-richer dynamics. 

\begin{figure}[ht]
  \centering
  \includegraphics[width=0.9\linewidth]{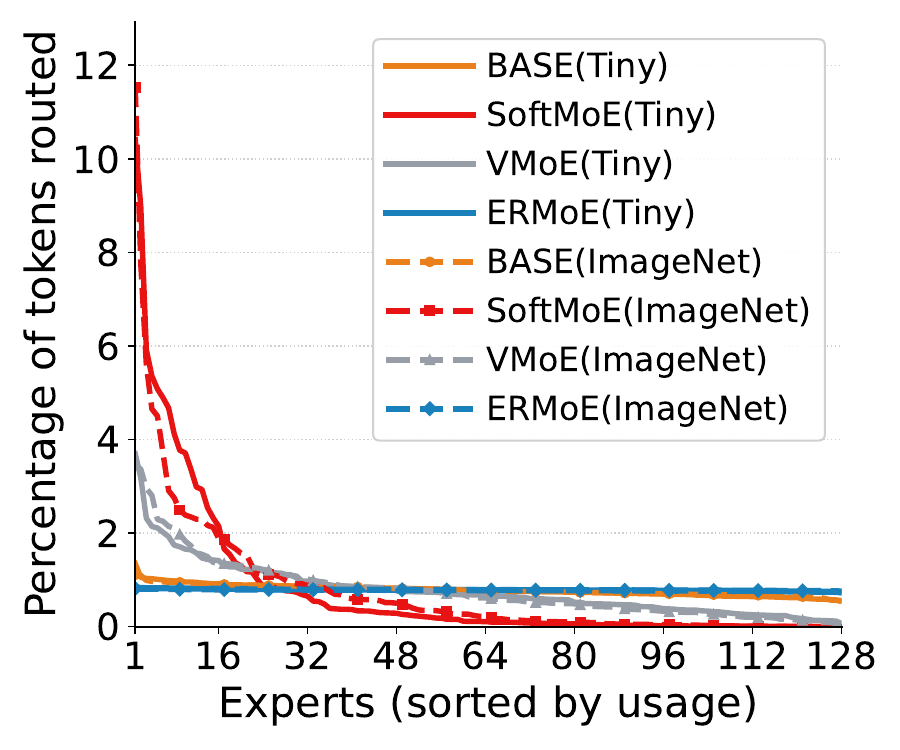}
  \caption{\textbf{Expert balancing on Tiny-ImageNet (test only) and ImageNet.} Percentage of tokens routed to each expert after sorting experts by usage (lower is flatter). We compare BASE (test-time greedy), SoftMoE (top-1 of soft combine weights), V-MoE (token-choice with LBL), and \textbf{ERMoE} (ours). ERMoE reduces the long-tail skew observed in V-MoE and the top-1 view of SoftMoE, approaching the near-uniform behavior of BASE without relying on balance-by-construction. Curves are computed on Tiny-ImageNet and ImageNet validation tokens with 128 total experts (16×8).}
  \label{fig:expert_comp}
\end{figure}

\textbf{Expert Balancing.} We conducted load balance analysis on both Tiny-ImageNet (val) and ImageNet (val) by sorting the 128 experts (16 layers$\times$8 experts per layer) by usage and plotting the percentage of test tokens routed to each expert (Fig.~\ref{fig:expert_comp}). Across both datasets, ERMoE yields a remarkably flatter curve than token-choice V-MoE~\cite{riquelme2021scaling}, and the top-1 projection of SoftMoE~\cite{puigcerver2023sparse}, indicating fewer stragglers and better utilization under test-time routing. In contrast, V-MoE, despite an auxiliary load-balancing loss, still shows a pronounced heavy tail, a behavior widely reported for token-choice MoEs. SoftMoE’s fully-differentiable mixing balances experts in aggregate, but when reduced to a discrete most-weighted expert for comparability, it exhibits a non-trivial tail that grows with depth. BASE~\cite{lewis2021base} serves as a reference: its layer solves a balanced assignment during training, and even the test-time greedy variant remains close to uniform, illustrating how balance-by-construction suppresses skew. Taken together, ERMoE approaches the near-uniform usage of BASE on both Tiny-ImageNet and ImageNet while avoiding auxiliary balancing losses and preserving content-aware routing, whereas conventional token-choice routers remain significantly imbalanced.

\textbf{Router selection for Brain Images} To test whether eigenbasis routing learns anatomically meaningful specialization, we examine region-conditioned selections over training. We fed the brain image with a specific region (WM/GM/CSF) to the model and then observed the expert activation status at the final MoE layer during different training stages (Visual examples are provided in the Appendix). As shown in Tab.~\ref {tab:router_region_selection}, at the early stage of training (epoch 1), the router does not know which expert to choose, so selections are diffuse and dominated by free experts with low scores. Because few scores exceed 0.5 early on, the router frequently engages the top-k fallback, yielding low-confidence, broadly shared assignments, which is exactly the behavior expected before expert eigenbases have specialized. With the training stage progressing to a deeper level (Epoch 300), routing has sharpened into anatomically meaningful primary choices, characterized by high eigenbasis scores. For example, csf\_expert takes over the input CSF image and yields a confident score of 0.91. Because we set top-k as 2, secondary selections remain, but at smaller weights. For CSF images, gm\_expert still shows some engagement reflecting plausible cross-region dependencies rather than collapse. Overall, the early$\rightarrow$late transition evidences a shift from fallback-driven, low-confidence assignments to stable, tissue-appropriate specialization with light, interpretable spillover.

\begin{table}[t]
\centering
\setlength{\tabcolsep}{10pt}
\renewcommand{\arraystretch}{1.05}
\begin{tabular}{lcc}
\toprule
\textbf{Region} & \textbf{Epoch 1} & \textbf{Epoch 300} \\
\midrule
\multirow{2}{*}{WM}
  & free\_expert1 (0.21) & wm\_expert (0.87) \\
  & free\_expert3 (0.13) & free\_expert1 (0.21) \\
\midrule
\multirow{2}{*}{GM}
  & free\_expert0 (0.15) & gm\_expert (0.79) \\
  & free\_expert1 (0.13) & wm\_expert (0.54) \\
\midrule
\multirow{2}{*}{CSF}
  & free\_expert4 (0.17) & csf\_expert (0.91) \\
  & free\_expert2 (0.17) & gm\_expert (0.47) \\
\bottomrule
\end{tabular}
\caption{\textbf{Router selection by region over training.} For each region (WM/GM/CSF), we report the top two experts with their eigenbasis scores at early (epoch~1) and late (epoch~300) stages.}
\label{tab:router_region_selection}
\vspace{-5mm}
\end{table}

%% file: sec/5_conclusion.tex
\section{Conclusion}
\label{sec:con}
\vspace{-3mm}
We introduce ERMoE, a sparse MoE that reparameterizes each expert in a learned orthonormal eigenbasis and routes tokens via eigen-alignment under a thresholded top-k. By tying assignment to representational subspaces, ERMoE removes auxiliary load-balancing losses and avoids the accuracy–stability trade-offs of learned-logit gates. On natural-image benchmarks, ERMoE matches state-of-the-art accuracy while approaching balance-by-construction, and it yields stronger few-shot linear probes—evidence of reduced redundancy and more linearly separable features. In neuroimaging, ERMoE-\emph{ba} attains state-of-the-art brain-age estimation on ADNI with anatomically meaningful expert usage. Looking ahead, we will extend to more cohorts to stress-test generalization; analyze eigenbasis formation and its link to expert diversity; and explore cross-modal variants that share eigenbases across imaging and clinical/tabular streams—pushing MoE systems toward models that are more accurate, better balanced, and easier to deploy in large-scale vision and clinical pipelines.

%% file: sec/6_acknowledgment.tex
\section*{Acknowledgment}
The authors A.C., S.D., S.L., C.Y., M.C., H.P., S.N., and P.B. acknowledge the support by the National Science Foundation (NSF) under the NSF Award 2243104 under the Center for Complex Particle Systems (COMPASS), the NSF Mid-Career Advancement Award BCS-2527046, the U.S. Army Research Office (ARO) under Grant No. W911NF-23-1-0111, the National Institute of Health (NIH) R01 AG 079957 "Interpretable machine learning to synergize brain age estimation and neuroimaging genetics", the Defense Advanced Research Projects Agency (DARPA) Young Faculty Award and DARPA Director Fellowship Award under Grant Number N66001-17-1-4044, Intel faculty awards, Northrop Grumman grant, and the NIH grants R01 AG 079957  and RF1 AG 082201 "Neurovascular calcification and ADRD in two nonindustrial Native American populations". It was a wonderful experience designing and writing the grant application entitled "Neurovascular calcification and ADRD in two nonindustrial Native American populations" and awarded under RF1 AG 082201. The views, opinions, and/or findings in this article are those of the authors and should not be interpreted as official views or policies of the Department of War, the National Institute of Health or the National Science Foundation.

%% file: sec/X_suppl.tex
\clearpage
\setcounter{page}{1}
\setcounter{section}{0}
\setcounter{figure}{0}
\setcounter{table}{0}
\renewcommand{\thefigure}{S\arabic{figure}}
\renewcommand{\thetable}{S\arabic{table}}
\maketitlesupplementary

\section{Data and Code Availability}
Codes for ERMoE are publicly available at \url{https://github.com/Belis0811/ERMoE}. MRI data are from the public cohort ADNI. No relevant accession codes are required to access these data, and the authors had no special access privileges that others would not have to the data obtained from any of these databases.

\section{Training Algorithms}

\subsection{Training ERMoE}

We insert ERMoE layers into a ViT backbone by replacing FFNs at fixed blocks and training end-to-end with standard ViT optimization (AdamW, cosine decay), following the patch-embedding and MHSA design in the original ViT paper~\cite{dosovitskiy2020image}. Unlike token-choice routers in sparse ViT variants (e.g., V-MoE) that learn free gating logits and then rely on auxiliary balancing losses, our router computes a cosine score in each expert’s learned eigenbasis and selects experts via a thresholded top-$k$ rule (Alg.~\ref{alg:ermoe}). This ties routing decisions directly to an expert’s representational subspace and obviates auxiliary load-balancing losses that prior work found necessary for stability in MoE layers~\cite{riquelme2021scaling,fedus2022switch}. An orthogonality penalty on the expert bases conditions the representation and prevents collapse, and the (rare) fallback path guarantees progress when no score exceeds the threshold.

\begin{algorithm}[ht]
\DontPrintSemicolon
\caption{Training ERMoE (thresholded top-$k$ with eigenbasis-aligned routing}
\label{alg:ermoe}
\SetKwInOut{Input}{Input}\SetKwInOut{Params}{Hyperparams}\SetKwInOut{Output}{Output}

\Input{Dataset $\mathcal{D}=\{(x,y)\}$; ViT backbone with $M$ MoE blocks; experts per block $E$}
\Params{top-$k$; threshold $T$; orthogonality weight $\lambda$; optimizer $\mathcal{O}$}
\Output{Trained parameters $\Theta$ (backbone + expert bases/coeffs)}

\BlankLine
\textbf{Initialize:} For each expert $e\in\{1,\dots,E\}$, initialize an orthonormal basis $B_e$ (columns) and expert parameters $\Phi_e$.\;
\BlankLine

\For{each minibatch $\mathcal{B}\subset\mathcal{D}$}{
  Tokenize every image $x\in\mathcal{B}$ into patches; embed to tokens $\{t_i\in\mathbb{R}^d\}$ with positional encodings.\;

  \For{$\ell\gets 1$ \KwTo $M$}{
    Apply MHSA and residual connections (ViT block) to obtain token features $\{z_i\}$ and attention-weighted contexts $\{\overline{c}_i\}$.\;

    \tcp{ERMoE routing within block $\ell$}
    \For{each token $z_i$ (and its context $\overline{c}_i$)}{
      \For{each expert $e\in\{1,\dots,E\}$}{
        $\tilde{x}_i \leftarrow x_i / \|x_i\|_2$;\quad $\tilde{c}_i \leftarrow \overline{c}_i / \|\overline{c}_i\|_2$ \tcp*{Normalize}
        $u_{i}^{(e)}\leftarrow B_e^\top \tilde{x}_i$ \tcp*{Project token}
        $v_{i}^{(e)}\leftarrow B_e^\top \tilde{c}_i$ \tcp*{Project context}
        $s_{i,e}\leftarrow \cos\!\big(u_{i}^{(e)},\,v_{i}^{(e)}\big)$ \tcp*{eigenbasis score}
      }
      $\mathcal{E}_i \leftarrow \{\,e \mid s_{i,e}>T\,\}$\;
      \eIf{$|\mathcal{E}_i|\ge k$}{
        $\mathcal{S}_i \leftarrow \text{top-}k\text{ experts in }\mathcal{E}_i\text{ by }s_{i,e}$\;
      }{
        $\mathcal{S}_i \leftarrow \text{top-}k\text{ experts over }\{1,\dots,E\}\text{ by }s_{i,e}$ \tcp*{fallback if no/too few exceed $T$}
      }
      Set $w_{i,e}\propto s_{i,e}$ for $e\in\mathcal{S}_i$; set $w_{i,e}=0$ otherwise; normalize $\sum_e w_{i,e}=1$.\;
      Compute expert outputs $h_{i,e}\leftarrow \mathrm{Expert}_e(z_i;\Phi_e)$ for $e\in\mathcal{S}_i$.\;
      Combine $y_i \leftarrow \sum_{e\in\mathcal{S}_i} w_{i,e}\,h_{i,e}$.\;
    }
    Apply residual and normalization to form the block output.\;
  }

  \textbf{Task loss:} $L_{\text{task}}$ (e.g., cross-entropy).\;
  \textbf{Orthogonality penalty:} $L_{\text{orth}}=\sum_{e}\big\|B_e^\top B_e-I\big\|_F^2$.\;
  Total loss $L \leftarrow L_{\text{task}}+\lambda L_{\text{orth}}$.\;
  Update $\Theta$ using optimizer $\mathcal{O}$.\;
}
\end{algorithm}

\subsection{Training ERMoE-\textit{ba}}

For 3D brain MRI, we adopt a 3D ViT tokenizer (fixed $16^3$ cubes) and instantiate a bank of experts that mixes anatomically targeted “region experts” (WM/GM/CSF) with “free experts”. Region experts can be warm-started using volumes in which the region of interest is retained (parcellations derived from standard neuroimaging tools, such as FreeSurfer, are a common way to isolate tissue classes). The same eigenbasis score and thresholded top-$k$ gate are used at every MoE block, and a lightweight regression head predicts brain age from the [CLS] token(Alg.~\ref{alg:ermoe_ba}). This design preserves the content-aware routing of ERMoE while accommodating volumetric context and region-specific specialization.

\begin{algorithm}[ht]
\DontPrintSemicolon
\caption{Training ERMoE-\textit{ba} for brain-age prediction (3D ViT with region \& free experts)}
\label{alg:ermoe_ba}
\SetKwInOut{Input}{Input}\SetKwInOut{Params}{Hyperparams}\SetKwInOut{Output}{Output}

\Input{T1 volumes $\mathcal{D}=\{(V,\mathrm{age})\}$; 3D ViT backbone with $M$ MoE blocks}
\Params{experts per block $E$; top-$k$; threshold $T$; orthogonality weight $\lambda$; BA head $g(\cdot)$; optimizer $\mathcal{O}$}
\Output{Trained parameters $\Theta_{\text{3D}}$}

\BlankLine
\textbf{Expert sets:} region experts $\mathcal{R}=\{\texttt{wm},\texttt{gm},\texttt{csf}\}$ (optionally warm-started on region-isolated inputs); free experts $\mathcal{F}$ (random init). Initialize $(B_e,\Phi_e)$ for all $e\in\mathcal{R}\cup\mathcal{F}$.\;

\BlankLine
\For{each minibatch $\mathcal{B}\subset\mathcal{D}$}{
  Tokenize each $V$ into non-overlapping $16{\times}16{\times}16$ cubes; embed tokens $\{t_i\in\mathbb{R}^d\}$ with 3D positional encodings.\;

  \For{$\ell\gets 1$ \KwTo $M$}{
    Apply 3D MHSA and residual connections to obtain $\{z_i\}$.\;
    \For{each token $z_i$}{
      \For{each expert $e\in\{1,\dots,E\}$}{
        $\hat{u}_{i,e}\leftarrow \mathrm{norm}(B_e^\top z_i)$;\quad
        $\hat{v}_e\leftarrow \mathrm{norm}(\psi_e)$;\quad
        $s_{i,e}\leftarrow \cos\!\big(\hat{u}_{i,e},\hat{v}_e\big)$.\;
      }
      $\mathcal{E}_i\leftarrow\{e\mid s_{i,e}>T\}$;\quad
      $\mathcal{S}_i\leftarrow\text{top-}k(\mathcal{E}_i)$ if $|\mathcal{E}_i|\ge k$, else top-$k$ overall.\;
      $w_{i,e}\propto s_{i,e}$ for $e\in\mathcal{S}_i$; normalize $\sum_e w_{i,e}=1$.\;
      $h_{i,e}\leftarrow\mathrm{Expert}_e(z_i;\Phi_e)$ for $e\in\mathcal{S}_i$;\quad
      $y_i\leftarrow\sum_{e\in\mathcal{S}_i} w_{i,e}\,h_{i,e}$.\;
    }
  }
  \textbf{Brain-age head:} $\hat{\mathrm{BA}}\leftarrow g\!\big(\texttt{[CLS]}(\{y_i\})\big)$.\;
  \textbf{Task loss:} $L_{\text{MAE}}=|\hat{\mathrm{BA}}-\mathrm{age}|$.\;
  \textbf{Orthogonality penalty:} $L_{\text{orth}}=\sum_e\|B_e^\top B_e-I\|_F^2$.\;
  Total loss $L \leftarrow L_{\text{MAE}}+\lambda L_{\text{orth}}$;\quad update $\Theta_{\text{3D}}$ with $\mathcal{O}$.\;
}
\end{algorithm}

Our procedures are compatible with common sparse-MoE infrastructure. ERMoE slots into ViT exactly where prior sparse FFN-replacements do (e.g., V-MoE), but removes the need for auxiliary load-balancing terms and router-specific tricks often required for stable training at scale.

\section{Visualization for brain router selection}
\begin{figure*}[t]
  \centering
  \includegraphics[width=\linewidth]{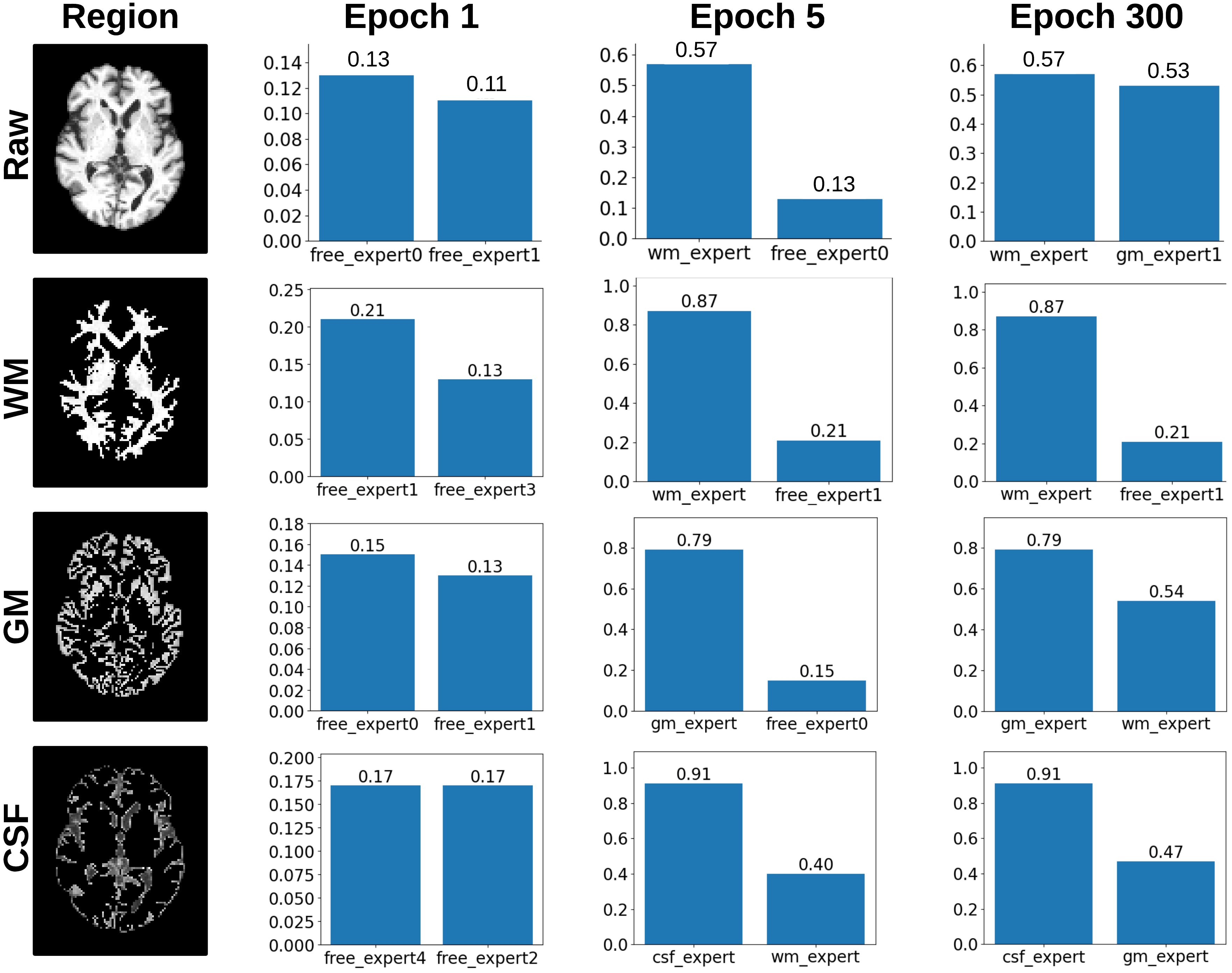}
  \caption{\textbf{Router selections on region-isolated brain inputs.} For each region (WM, GM, CSF), we feed a volume in which only that region is retained and log the experts chosen by the final MoE layer over training. Columns show epochs 1, 5, and 300; each cell lists the top–2 selected experts with their eigenbasis scores.}

  \label{fig:visual_brain}
\end{figure*}

To probe whether ERMoE-\textit{ba} learns anatomically meaningful specializations, we visualize router behavior with \emph{region-isolated} inputs. For each tissue—white matter (WM), gray matter (GM), and cerebrospinal fluid (CSF)—we retain only that region in the volume (visual exemplars in the Appendix) and record the top-2 experts selected by the final MoE layer along with their eigenbasis scores at three training checkpoints (epochs 1, 5, and 300), as shown in Fig.~\ref{fig:visual_brain}. WM/GM/CSF masks are derived from a standard neuroimaging pipeline (FreeSurfer), ensuring that the inputs reflect well-established tissue contrasts.

At early training (epoch 1), selections are diffuse and dominated by free experts with modest eigenbasis scores, consistent with immature expert subspaces. By epoch 300, routing sharpens into tissue-appropriate specializations: \texttt{wm\_expert} is the primary choice for WM, \texttt{gm\_expert} dominates GM, and \texttt{csf\_expert} dominates CSF. Secondary choices (e.g., \texttt{wm\_expert} for GM at 0.54; \texttt{gm\_expert} for CSF at 0.47) persist at lower weights, reflecting plausible cross-tissue dependencies rather than collapse. The emergence of such structured, content-aware routing mirrors observations in vision MoEs, where deeper layers exhibit stronger alignment between routing decisions and semantic categories, but here the categories are anatomical tissues rather than object classes.

Methodologically, this region-isolation analysis serves as a targeted perturbation test: by ablating all but one tissue, we assess whether the router’s selections remain stable and interpretable under controlled input changes—akin in spirit to occlusion/meaningful-perturbation probes widely used in interpretability. The consolidation of high-confidence selections by epoch 300 indicates that ERMoE’s eigenbasis-guided routing learns anatomically grounded expert subspaces rather than relying on spurious correlations.

\section{More Results on ADNI}

\begin{figure}[t]
  \centering
  \includegraphics[width=\linewidth]{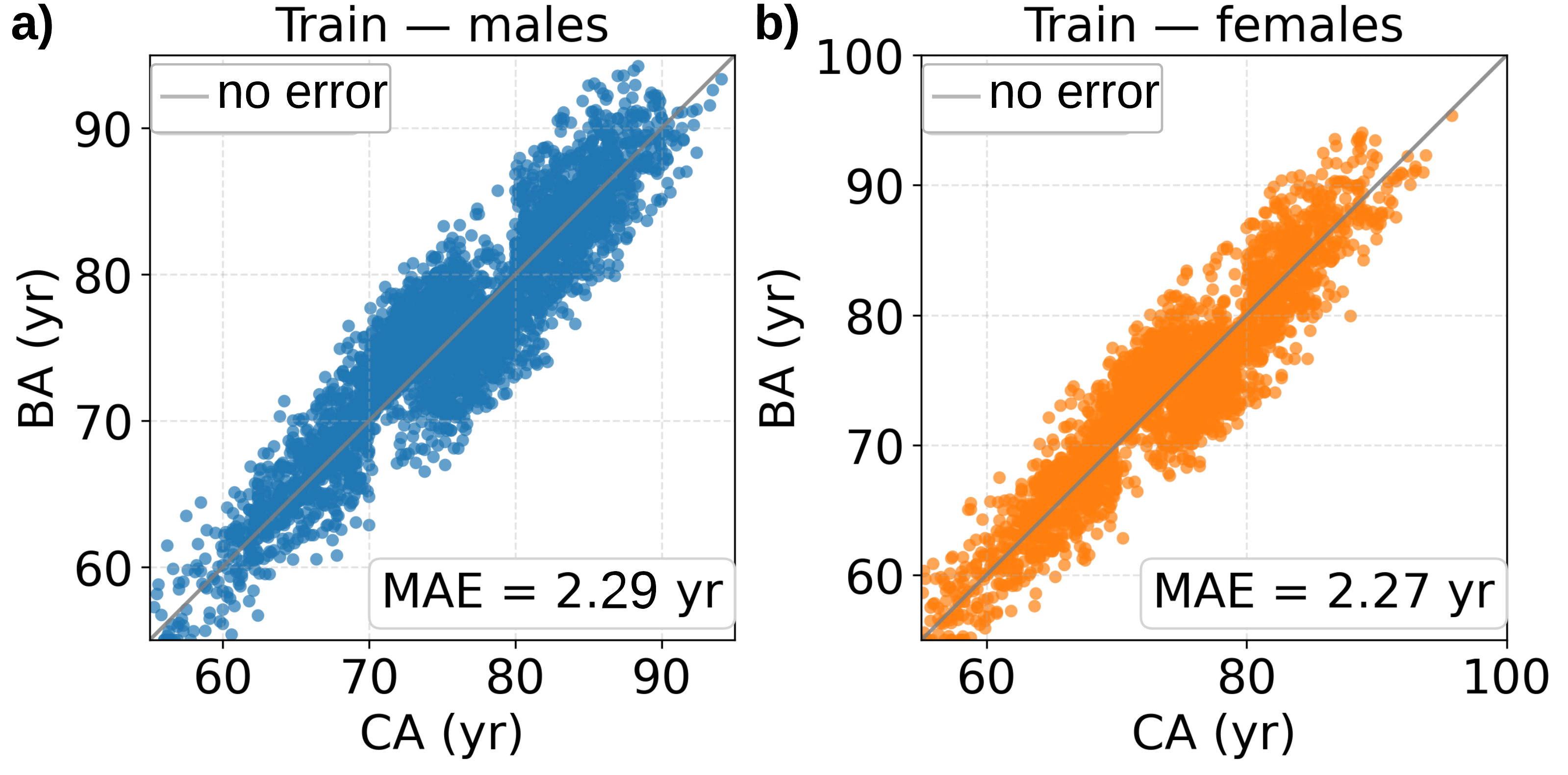}
  \caption{\textbf{Brain-age (BA) estimation on ADNI train set.}
Scatter plots show ERMoE-\textit{ba} predicted BA versus chronological age (CA) for a) males and b) females on the test set. Points are colored by sex (male: blue; female: orange). The solid diagonal denotes the ``no error'' line ($\mathrm{BA}=\mathrm{CA}$).}
  \label{fig:mae_train}
  \vspace{-7mm}
\end{figure}

We examined ERMoE-\textit{ba} on the ADNI \emph{train} set to verify that the model learns a stable and well-calibrated mapping from chronological age (CA) to predicted brain age (BA). Figure~\ref{fig:mae_train} shows sex-stratified BA–CA scatter plots for males (panel a) and females (panel b). In both groups, points cluster tightly around the identity line (BA = CA), with only modest dispersion across the full age range. The resulting train MAEs are \textbf{2.29} years for males and \textbf{2.27} years for females, which are close to the corresponding test MAEs reported in Table~\ref{tab:adni_ba_mae}. 

The near-unit slope and small intercept of the best-fit trends, together with the absence of obvious curvature at younger or older ages, suggest that ERMoE-\textit{ba} attains good calibration on ADNI without relying on aggressive regularization or post-hoc correction. Moreover, the similar spread between male and female panels indicates that the router learns age-sensitive morphometric patterns that generalize across sex, rather than overfitting to sex-specific shortcuts. These training results support that ERMoE-\textit{ba} fits ADNI without evident overfitting and maintains consistent calibration across demographic subgroups.

\section{Post-hoc age-level calibration}
Brain-age residuals (BA–CA) typically exhibit age dependence rather than Independent and Identically Distributed(IID), mean-zero Gaussian noise. Following best practice, we evaluate a simple post-hoc linear calibration learned on the training split only: we fit $\hat{y}=a+by$ on the training data and correct test predictions via $\hat{y}_{corr}=(\hat{y}-a)/b$. As shown in Tab.~\ref{tab:posthoc}, we report MAE, $corr(\hat{y}-y,y)$, and the calibration slope/intercept from regressing $\hat{y}$ on $y$. This addresses the characteristic “regression-to-the-mean” bend without inflating metrics, aligning with recommendations to report both raw and bias-corrected results, and avoiding over-corrections that artificially boost accuracy.

On the ADNI held-out test set (n=$797$; M=$460$, F=$337$), pooled calibration reduces age-dependence (corr from $-0.287$ to $-0.125$) and improves calibration (slope $0.887$ to $0.949$) with a small MAE gain ($2.307$ to $2.287$ years). Sex-specific calibration yields nearly identical calibration and no further MAE benefit ($2.295$ y). We therefore report both raw and pooled-calibrated metrics in the main text. Recent age-level bias work further motivates reporting such diagnostics alongside primary accuracy.

\begin{table}[H]
\center
\small
\begin{tabular}{lcccc}
\toprule
Method & MAE& corr & Cal. slope & Cal. intercept \\
\midrule
Raw & \textbf{$2.307$} & $-0.287$ & $0.887$ & $9.026$ \\
pooled & \textbf{$2.287$} & $-0.125$ & $0.949$ & $4.221$ \\
sex-specific & \textbf{$2.295$} & $-0.126$ & $0.949$ & $4.244$ \\
\bottomrule
\end{tabular}
\caption{\textbf{Age-level calibration ablation on ADNI test.} Calibration parameters are learned on the training split only (no leakage).}
\label{tab:posthoc}
\end{table}

\section{Different Thresholds} 
\begin{figure}[ht]
    \centering
    \includegraphics[width=\linewidth]{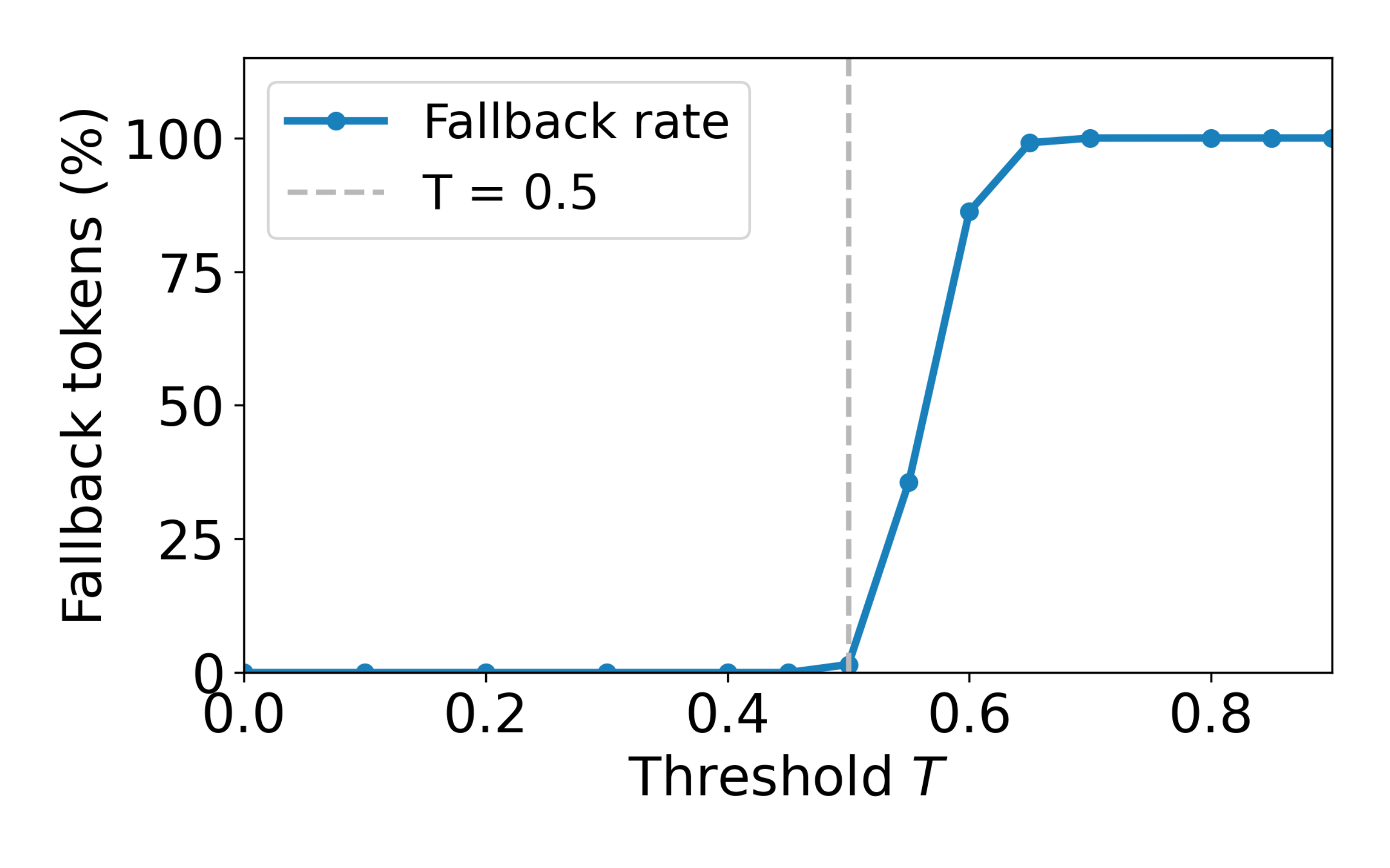}
    \caption{\textbf{Threshold $T$ controls fallback in ERMoE (Tiny\mbox{-}ImageNet, test only).} A fallback occurs only when no expert’s eigenbasis score exceeds $T$, in which case the router reverts to global top-$k$ ($k{=}2$). The curve stays near zero for $T{<}0.5$ and rises sharply for $T{\ge}0.5$, indicating that $T{=}0.5$ prunes noisy routes without starving eligibility (lower is better).}
    \label{fig:fallback_rate}
\end{figure}

To show the effectiveness of our router, we ablate the router threshold $T$ that filters experts by eigenbasis score before top-$k$ selection. At test time, we freeze all weights and sweep $T\!\in\![0,0.9]$ on Tiny-ImageNet (val), counting a fallback whenever no expert satisfies $\max_e\mathrm{score}_e(x)\!>\!T$, in which case the router reverts to global top-$k$. The fallback curve (Fig.~\ref{fig:fallback_rate}) remains near zero for $T{<}0.5$ and rises once $T{\ge}0.5$. This shows that when we set a low $T$, the model will admit low-confidence experts and reintroduce the noisy, content-misaligned assignments characteristic of token-choice routing. This willsignificantly affect accuracy and the utilization of expertsn. When $T$ is too high(higher than $0.5$), the router will fall back to global top-$k$ selections, which overloads a small subset of experts and recreates long-tail utilization.. Based on this behavior, we fix $T\!=\!0.5$ for all main results.

\section{Choosing the Orthogonality Weight.} As defined in our loss function (Eq.~\ref{eq:loss}), $\lambda$ trades off the classification term $L_{\mathrm{CLS}}$ and the orthogonality regularizer $L_{\mathrm{orth}}$ that encourages the expert bases to remain (near-)orthonormal. Orthogonality/spectral penalties are known to reduce feature redundancy and stabilize optimization in deep networks, but they must be applied with \emph{small} coefficients to avoid overwhelming the task objective~\cite{yoshida2017spectral,bansal1810can}. 
We therefore sweep $\lambda \!\in\! [0, 0.01]$ and train for 15 epochs on a 10\% ImageNet dataset to locate a robust operating point.

\begin{figure}[t]
    \centering
    \includegraphics[width=.8\linewidth]{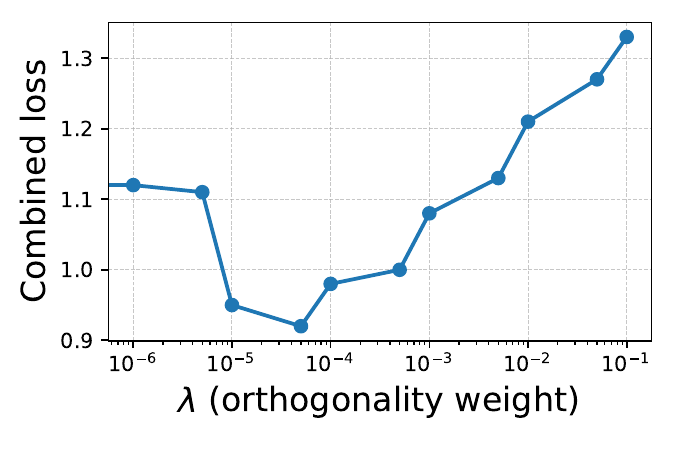}
    \caption{\textbf{Effect of the orthogonality weight $\lambda$.}
    Minimum validation loss is achieved at $\lambda=5\times10^{-5}$.}
    \label{fig:lambda_orth}
    \vspace{-5mm}
\end{figure}

As shown in Fig.~\ref{fig:lambda_orth}, the combined validation loss follows a clear U-shape with a minimum at $\lambda=5\times10^{-5}$. 
For very small values ($\leq10^{-5}$), the penalty under-regularizes, yielding poorly conditioned bases and slower convergence; for large values ($\geq 10^{-3}$), the penalty dominates and degrades discriminative learning—consistent with prior observations on orthogonality regularization strength. 
Accordingly, we \textbf{fix $\lambda=5\times10^{-5}$} in all subsequent experiments.

\section{Effectiveness of using top-k}
We fix the routing threshold at $T{=}0.5$ and vary the number of selected experts $k\!\in\!\{1,2,3,4,6\}$ in each ERMoE layer. For each $k$, we evaluate three quantities on Tiny-ImageNet validation set: (i) task accuracy (Top-1), (ii) routing \emph{tail mass}, and (iii) normalized step time. Tail mass measures how much probability leaks into marginal experts that merely clear the threshold and would be pruned by a smaller $k$. Formally, with scores $\{s_e\}$, set $A\!=\!\{e:\,s_e>T\}$ and let $S_k\!\subset\!A$ index the top-$k$ experts by score; with mixture weights $w_e\!\propto\!s_e$ renormalized on $A$,
\begin{equation}
\textstyle
\mathrm{TailMass}
\;=\;
\frac{\sum_{e\in A\setminus S_k} w_e}{\sum_{e\in A} w_e}.
\label{eq:tailmass}
\end{equation}

\begin{figure}[ht]
  \centering
  \includegraphics[width=\linewidth]{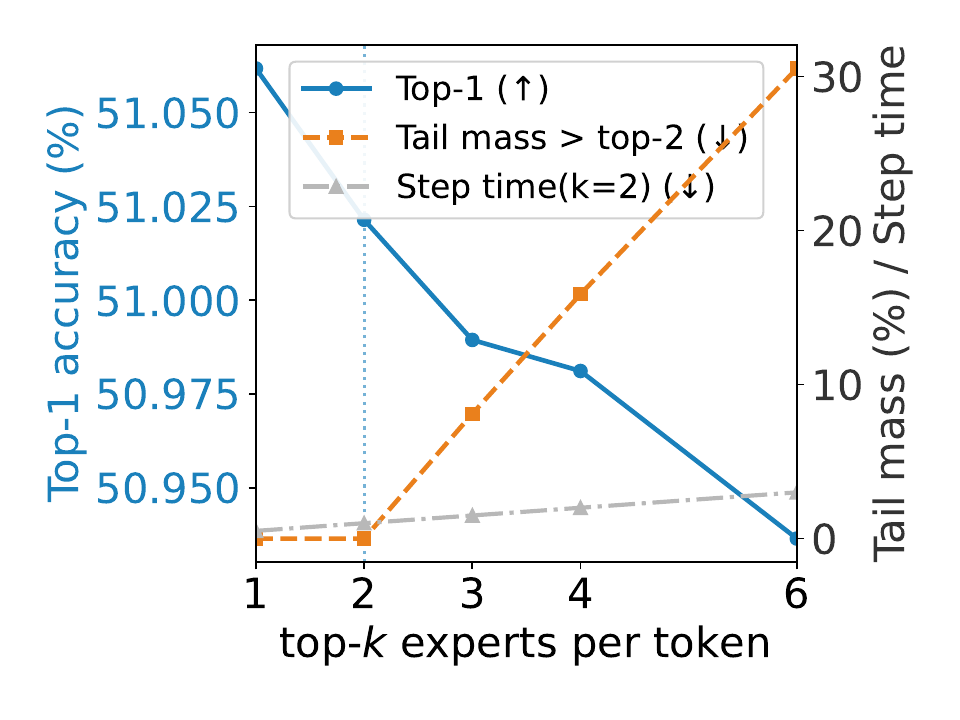}
  \caption{\textbf{Top-$k$ ablation at fixed threshold $T{=}0.5$ on Tiny-ImageNet (val).}
  We vary the number of selected experts per token $k\!\in\!\{1,2,3,4,6\}$ and report Top-1 (left axis), routing tail mass from Eq.~\eqref{eq:tailmass} (right axis), and normalized step time. 
  $k{=}1$ underutilizes capacity; $k{=}2$ yields the best accuracy with a low tail mass and moderate cost; $k{>}2$ increases tail mass and latency while reducing accuracy due to mixture dilution by low-scoring, above-threshold experts.}
  \label{fig:topk_ablation}
\end{figure}

Figure~\ref{fig:topk_ablation} shows a consistent pattern. \textbf{(1) $k{=}1$ under-selects capacity.} Accuracy drops because many tokens that meet the threshold still benefit from a secondary expert; restricting to a single expert sacrifices complementary subspaces. \textbf{(2) $k{=}2$ is the sweet spot.} Accuracy peaks while tail mass remains low, indicating that most of the useful probability mass concentrates on two high-confidence experts; latency remains near the sparse MoE budget. \textbf{(3) $k{>}2$ dilutes predictions and slows inference.} As $k$ grows, more above-threshold but low-score experts enter the mixture, tail mass increases monotonically, and Top-1 degrades due to averaging in weak directions; step time rises roughly linearly with the number of active experts per token. In short, given a fixed content threshold $T$, \emph{top-$k$ is necessary} to cap marginal contributors, preserve sparsity, and avoid accuracy loss from mixture dilution. This finding aligns with observations in the sparse MoE literature that controlling the number of active experts is critical for both computational efficiency and quality. 

\section{Training Efficiency.} 

Another key aspect for MoE architectures is scaling model capacity while controlling computational cost. We benchmark efficiency at 224$^2$ with ViT\textendash B/16 and $E{=}8$ experts. As shown in table ~\ref{tab:efficiency}, compared to V\text{-}MoE, \textbf{ERMoE} reduces computation while keeping the same backbone and gating budget: FLOPs dropped \textbf{10.3\%}, and inference time falls from 4.4,ms to \textbf{4.1,ms}. Total trainable parameters shrink \textbf{14.7\%}, and active parameters per sample showed \textbf{12.9\%} reduction. The “active parameters” notion is standard in sparse MoE: only the routed experts participate in a given forward/backward pass, so reductions here translate directly into lower per-step math and memory traffic.

\begin{table}[H]
\centering
\small
\setlength{\tabcolsep}{4pt}
\caption{\textbf{Compute and efficiency comparison.} “Params (M)” shows total parameters; “(act)” shows per-sample \emph{active} parameters. FLOPs are for 224$\times$224. V-MoE numbers are for ViT-B/16 with 8 experts, top-2 gating every 2 blocks. Single-gated MoE and DeepMoE report on ResNet-18 (largest reported CNN settings).}
\begin{tabular}{lcccc}
\toprule
Model  & Params (M) & FLOPs (G) & Inference \\
\midrule
Single-gated MoE & n/r & 2.18 & n/r  \\
DeepMoE  & n/r & 1.81 & n/r  \\
V-MoE   & 284.9 \,(129.9 act) & 26.3 & 4.4 ms\\
\midrule
\textbf{ERMoE} & 243.0 \,(113.2 act) & 23.6 & 4.1 ms \\
\bottomrule
\end{tabular}
\label{tab:efficiency}
\vspace{-5mm}
\end{table}

These results confirms that ERMoE achieved superior accuracy without introducing any more trainable parameters or consuming more computational resources. 

\section{More ablations.} 

To address this directly, we added component ablations on CIFAR-10 5-shot classification, keeping experts($E$), top-k, and $T$ fixed to default settings. We report 3 variants (i)with standard expert weights, (ii)learned routing logits, and (iii)no orthogonality($\lambda = 0$). 
Tab.\ref{tab:ablation} shows that replacing eigenbasis experts with standard weights drops accuracy by 2\%, indicating the parameterization is more than compression and strengthens the intended router--expert subspace coupling. Replacing cosine eigenbasis scoring yields the largest drop (5\%) and a much higher CV$^2$(squared coefficient of variation), proving that token-choice learned routers often needing auxiliary balancing to avoid heavy-tail expert usage. Setting $\lambda{=}0$ has little effect on accuracy but increases expert subspace overlap(inclined CV$^2$ and Fallback\%).
\vspace{-3mm}
\begin{table}[H]
\centering
\small
\setlength{\tabcolsep}{2pt}
\begin{tabular}{lccc}
\toprule
Variant & Top-1 (\%) $\uparrow$ & Load CV$^2$ $\downarrow$ & Fallback (\%) $\downarrow$ \\
\midrule
normal weights  & 94.12 & 0.10 & 0.3 \\
learned-logit router & 91.07  & 0.32 & 2.4 \\
$\lambda{=}0$ & 95.35 & 0.15 & 0.7 \\
\midrule
ERMoE  & \textbf{96.05}  & \textbf{0.08} & \textbf{0.2} \\
\bottomrule
\end{tabular}
\vspace{-3mm}
\caption{\textbf{Component ablations.} Load CV$^2$ computed over per-expert token counts. Fallback compute same as Appendix Sec.6}
\label{tab:ablation}
\vspace{-5mm}
\end{table}

Overall, the ablations show that cosine alignment is the main source of the gains; eigen-parameterized experts add a consistent improvement; orthogonality mostly improves stability rather than being the sole driver of accuracy.

\section{Routing Overhead and Practical Runtime}

ERMoE routing computes an Eigenbasis Score by projecting each token feature (and its attention-weighted context) into each expert basis and then taking cosine similarity. Compared to a token-choice router that computes logits using a single linear map, this adds routing arithmetic. Table~\ref{tab:sup_routing_flops} reports routing-step FLOPs and routing-step wall-clock.

\begin{table}[t]
\centering
\scriptsize
\setlength{\tabcolsep}{6pt}
\begin{tabular}{lcc}
\toprule
Routing & FLOPs (G) & Wall-clock (ms)  \\
\midrule
V-MoE  & 0.631 & 0.03 \\
ERMoE & 1.169  & 0.05 \\
\bottomrule
\end{tabular}
\vspace{-2mm}
\caption{\textbf{Routing cost.} V-MoE denotes a standard vision MoE routing baseline.}
\label{tab:sup_routing_flops}
\end{table}

For scaling, the routing compute is linear in the number of experts:
\begin{equation}
\text{routing FLOPs} = O(B\,E\,d\,r),
\end{equation}
where $B$ is the number of tokens, $E$ is the number of experts, $d$ is feature dimension, and $r$ is the basis rank. The per-expert routing-state memory is also linear in $E$:
\begin{equation}
\text{routing-state memory} = O(E\,d\,r^2).
\end{equation}
Orthogonality is enforced mainly via $L_{\text{ortho}}$ during training, and we apply a lightweight re-orthonormalization to each expert basis once per epoch; the amortized cost is $O(E\,d\,r^2)$ per epoch.

\section{Scalability With Expert Count}

To provide scaling evidence under limited compute, we run a one-layer stress test at $E{=}8$ vs.\ $E{=}32$ and measure allocated GPU memory for one epoch under the same setting. ERMoE uses 20.5GB ($E{=}8$) and 84.7GB ($E{=}32$), while a standard MoE uses 18.0GB and 80.0GB, respectively. Thus, ERMoE adds 2.2GB while increasing $E$ by $4\times$.

Since ERMoE does not perform any online $d\times d$ eigendecomposition, the dominant large-$E$ bottleneck remains sparse MoE execution (token dispatch/gather and expert compute), rather than ERMoE scoring.

\section{Additional Baseline: Expert-Choice}

We add an Expert-Choice baseline on ImageNet under the same experimental settings. Expert-Choice achieves 78.95\% top-1 accuracy, while ERMoE reaches 88.03\% top-1 accuracy.

\section{Multimodal Setting: Fine-tuning the Text Tower}

We fine-tune the text tower in Fix-CLIP and evaluate text-to-image retrieval on COCO. The results are 50.3\% R@1, 74.0\% R@5, and 82.6\% R@10, improving over the frozen-text Fix-CLIP setting (49.1\%, 73.8\%, and 82.4\%). These results show that joint tuning of the text encoder can improve T2I retrieval, while our frozen-text experiments serve as a controlled setting to isolate the image encoder’s contribution.